\documentclass[conference]{IEEEtran}
\IEEEoverridecommandlockouts
\usepackage{cite}
\usepackage{amsmath,amssymb,amsfonts}
\usepackage{algorithmic}
\usepackage{graphicx}
\usepackage{textcomp}
\usepackage{xcolor}
\usepackage{enumitem}
\usepackage{cite}
\usepackage{hyperref}
\usepackage{adjustbox} 
\usepackage{float}

\usepackage{tabularx,booktabs}
\newcolumntype{C}{>{\centering\arraybackslash}X}
\def\BibTeX{{\rm B\kern-.05em{\sc i\kern-.025em b}\kern-.08em
    T\kern-.1667em\lower.7ex\hbox{E}\kern-.125emX}}

\begin{document}

\title{Mask-to-Height: A YOLOv11-Based Architecture for Joint Building Instance Segmentation and Height Classification from Satellite Imagery}


\author{
\IEEEauthorblockN{Mahmoud El Hussieni, Bahadır K. Güntürk}
\IEEEauthorblockA{Istanbul Medipol University\\ 34810,  Istanbul, Türkiye\\
mahmoud.moawed@std.medipol.edu.tr \\bkgunturk@medipol.edu.tr}
\and
\IEEEauthorblockN{Hasan F. Ateş} 
\IEEEauthorblockA{Dept. of AI and Data Eng.\\ Ozyegin University \\ 34794, Istanbul, Türkiye\\
hasan.ates@ozyegin.edu.tr}
\and
\IEEEauthorblockN{Oğuz Hanoğlu}
\IEEEauthorblockA{Huawei Türkiye R\&D Center \\ 34768, Istanbul, Türkiye\\
oguz.hanoglu1@huawei.com}
}

\IEEEoverridecommandlockouts
\IEEEpubid{\makebox[\columnwidth]{979-8-3315-9727-6/25/\$31.00~\copyright2025 IEEE \hfill}
\hspace{\columnsep}\makebox[\columnwidth]{ }}

\maketitle

\begin{abstract}
Accurate building instance segmentation and height classification are critical for urban planning, 3D city modeling, and infrastructure monitoring. This paper presents a detailed analysis of YOLOv11, the recent advancement in the YOLO series of deep learning models, focusing on its application to joint building extraction and discrete height classification from satellite imagery. YOLOv11 builds on the strengths of earlier YOLO models by introducing a more efficient architecture that better combines features at different scales, improves object localization accuracy, and enhances performance in complex urban scenes. Using the DFC2023 Track 2 dataset---which includes over 125,000 annotated buildings across 12 cities---we evaluate YOLOv11's performance using metrics such as precision, recall, F1 score, and mean average precision (mAP). Our findings demonstrate that YOLOv11 achieves strong instance segmentation performance with 60.4\% mAP@50 and 38.3\% mAP@50--95 while maintaining robust classification accuracy across five predefined height tiers. The model excels in handling occlusions, complex building shapes, and class imbalance, particularly for rare high-rise structures. Comparative analysis confirms that YOLOv11 outperforms earlier multitask frameworks in both detection accuracy and inference speed, making it well-suited for real-time, large-scale urban mapping. This research highlights YOLOv11's potential to advance semantic urban reconstruction through streamlined categorical height modeling, offering actionable insights for future developments in remote sensing and geospatial intelligence.
\end{abstract}


\section{Introduction} \label{sec:intro}
Urban planning, disaster response, and environmental monitoring increasingly rely on accurate geospatial intelligence derived from satellite imagery. A key challenge lies in extracting both spatial boundaries and vertical characteristics of built environments at scale.

This paper presents a unified framework for joint building instance segmentation and discrete height classification using the latest version of the You Only Look Once (YOLO) architecture—YOLOv11. Unlike regression-based methods that output continuous height values (e.g., 17m), we classify buildings into interpretable height categories (e.g., ``low-rise,'' ``high-rise''). This reframing offers significant advantages: it simplifies downstream applications such as zoning and infrastructure planning, enhances robustness to noisy or incomplete elevation data, and eliminates the need for complex post-processing.

Our approach leverages the DFC2023 Track 2 dataset, which consists of multimodal satellite imagery from 12 cities across five continents. It includes over 125{,}000 annotated buildings, with normalized Digital Surface Models (nDSMs) providing ground truth height information. Buildings are categorized into five height classes, as defined in Table~\ref{tab:height_class_definitions}.

We compare our method against recent state-of-the-art models including LIGHT~\cite{mao2023},  HGDNet~\cite{liu2023}, and the multitask network by Huo et al.~\cite{huo2022}. While these models perform continuous height regression and often rely on dense supervision and complex feature fusion, our categorical approach integrates height classification directly with instance segmentation, enabling more interpretable and deployment-friendly outputs.

Additionally, we demonstrate that YOLOv11’s real-time inference capabilities make it particularly suitable for large-scale urban mapping. By modeling height as a structured classification task, we improve interpretability, deployment efficiency, and resilience to label noise, while maintaining high spatial fidelity and detection accuracy.

The remainder of this paper is organized as follows: Section~\ref{sec:related_work} reviews related work; Section~\ref{sec:methodology} describes the dataset and methodology; Section~\ref{sec:results} presents experimental results and comparisons; Section~\ref{sec:conclusion} concludes the paper with a discussion of future directions.

\section{Related Work} \label{sec:related_work}
Recent advances in remote sensing and deep learning have led to the development of several multitask frameworks for joint building extraction and height estimation. One of the most notable approaches is LIGHT~\cite{mao2023}, which combines Mask R-CNN with a Pyramid Pooling Module (PPM) to perform pixel-wise height regression alongside instance segmentation. While LIGHT achieves competitive performance on the DFC2023 dataset, its reliance on continuous height prediction introduces complexity and sensitivity to noisy ground truth data. The model also employs a Gated Cross Task Interaction (GCTI) module to enhance feature sharing between tasks, further increasing architectural overhead.

Another prominent method is HGDNet~\cite{liu2023}, which uses hierarchical guided distillation to align semantic and geometric features across branches. This approach improves consistency between segmentation and height estimation but comes at the cost of increased training time and dependency on teacher networks. Similarly, Huo et al.~\cite{huo2022} proposed a multitask framework that jointly estimates building footprints and heights using shared backbone features. However, their method lacks the ability to distinguish individual building instances—a critical requirement for urban planning and 3D reconstruction applications.

Unlike prior works that rely on continuous height regression or external modules for task interaction, our framework adopts a categorical classification paradigm for building height modeling. By integrating instance segmentation and discrete height classification within a single, streamlined architecture—specifically YOLOv11—we avoid the need for complex multi-branch designs commonly used in models like LIGHT~\cite{mao2023} and HGDNet~\cite{liu2023}. Our approach achieves strong performance with 60.4\% mAP@50 and 38.3\% mAP@50--95 on average for five height classes, this indicates that discrete height classification can serve as an effective alternative to explicit regression in many cases. Moreover, the model supports real-time inference and deployment, making it well-suited for large-scale urban mapping tasks. These improvements highlight the practical advantages of discrete height modeling, particularly in the presence of class imbalance and measurement uncertainty typical of real-world remote sensing data indicating that discrete height classification can be an effective alternative to explicit regression in many cases.

\section{Methodology} \label{sec:methodology}
\subsection{Dataset and Height Classification Pipeline}
The experiments were conducted using the \textbf{IEEE GRSS Data Fusion Contest 2023 (DFC2023) Track 2} dataset~\cite{xia2023}, a large-scale benchmark designed to support semantic urban reconstruction by combining satellite imagery with normalized Digital Surface Models (nDSMs). The dataset comprises \textbf{1,773 multimodal satellite images} from 12 cities across five continents, featuring:

\begin{itemize}
    \item \textbf{Multimodal Satellite Imagery:} Includes both \textbf{optical orthophotos} (RGB channels) and \textbf{Synthetic Aperture Radar (SAR)} data.

    \item\textbf{Polygon annotations}: Precise vector outlines suitable for complex shapes and interior structures    
    \item \textbf{Normalized DSMs} used as ground truth for height values
\end{itemize}

A total of \textbf{125,153 annotated buildings} are included in the DFC2023 Track 2 dataset, with segmentation masks provided in polygon format. As defined in the challenge guidelines~\cite{xia2023}, Track 2 focuses on the joint task of \textbf{building extraction and continuous height estimation} from multimodal satellite imagery (optical and SAR). Participants are required to reconstruct building footprints and predict pixel-wise height values using Digital Surface Models (DSMs) as ground truth. 

In contrast to this regression-based objective, our approach adopts a classification-based formulation, a strategy that has shown effectiveness in various vision tasks.

We reformulate height estimation as a \textbf{discrete classification task}, assigning each building to one of five predefined height classes based on DSM-derived statistics as shown in Table~\ref{tab:height_class_definitions}. This approach offers greater robustness to noisy elevation data and aligns with practical urban planning needs, where coarse-grained tiers (e.g., 0–10m, 11–20m, etc.) are more actionable than precise values. The classes reflect typical zoning and building typologies, ranging from low-rise homes to skyscrapers, enabling more effective integration into applications like 3D city modeling, zoning regulation, and risk assessment \cite{stipek2024}.

\begin{table}[htbp]
\centering
\caption{Height Class Definitions Based on DSM Ranges}
\label{tab:height_class_definitions}
\begin{tabular}{cl}
\hline
\textbf{Class} & \textbf{Height Range (meters)} \\
\hline
1 & 0--10 \\
2 & 11--20 \\
3 & 21--30 \\
4 & 31--40 \\
5 & 41+ \\
\hline
\end{tabular}
\end{table}


\subsubsection{Digital Surface Model Processing and YOLO Annotation Generation}
To enable joint building instance segmentation and discrete height classification, we implemented a structured preprocessing pipeline that converts raw DSM data into YOLOv11-compatible annotations. This process follows established practices in remote sensing data preparation~\cite{xia2023} while adapting to the specific requirements of YOLOv11’s architecture~\cite{khanam2024}.

\begin{enumerate}

\item \textbf{Data Loading and Configuration}  
COCO-formatted annotations are loaded to extract building instances. Corresponding DSM rasters are read using the \texttt{rasterio} library for geospatial alignment. Output directories for YOLO-formatted labels are created to organize training and evaluation.

\item \textbf{Annotation Handling}  
Polygon annotations are prioritized for accurate boundary extraction. All coordinates are normalized to the [0,1] range required by YOLO format:
$$
x' = \frac{x}{W},\quad y' = \frac{y}{H}
$$
where $W$, $H$ are image width and height.\\

\item \textbf{Height Calculation}  
We convert continuous DSM values into five discrete height classes (Table~\ref{tab:height_class_definitions}), enabling interpretable, category-based height modeling that aligns with practical urban planning needs.

For each building instance, a binary mask is generated from its polygon, and DSM values within the masked area are extracted, with invalid (NaN) entries filtered out. A robust height estimate is then computed as the rounded mean of valid DSM values:
\begin{equation}
h_{\text{mean}} = \text{round}\left(\frac{1}{N}\sum_{i=1}^{N} h_i \right)
\label{eq:height_mean}
\end{equation}

where $h_i$ are valid DSM height samples within the building boundary.

\item \textbf{Height Class Assignment}  
Based on the mean height value calculated in the previous step, each building is assigned to one of the five predefined height classes, consistent with our discrete modeling approach.

\item \textbf{YOLO-Compatible Label Generation}  
The final output is stored in the standard YOLO format:

\vspace{0.4cm}
$\begin{aligned}
&\texttt{<class> <x1> <y1> <x2> <y2>...<xn> <yn>} \\
&\quad \text{(normalized polygon coordinates)}
\end{aligned}$

\end{enumerate}

This format enables seamless integration with YOLOv11’s instance segmentation and classification heads.

\subsubsection{Class Distribution and Balancing}
The dataset exhibits natural class imbalance, with lower-rise buildings dominating:
    
\begin{itemize}
    \item \textbf{Class 1 (0--10 m)}: 35.4\% 
    \item \textbf{Class 2 (11--20 m)}: 28.0\% 
    \item \textbf{Class 3 (21--30 m)}: 23.8\% 
    \item \textbf{Class 4 (31--40 m)}: 9.8\% 
    \item \textbf{Class 5 (41+ m)}: 3.1\% 
\end{itemize}

To mitigate this imbalance during training, we employed focal loss and adaptive class weighting strategies, preventing bias toward the majority classes while maintaining sensitivity to rare, taller building types.

     

The DFC2023 Track 2 dataset offers a large-scale, multimodal benchmark with precisely annotated buildings, combining spatial, elevation, and polygon-level data. Its complexity and inherent height class imbalance make it well-suited for evaluating multitask models that unify instance detection with structured categorical outputs—aligning closely with our approach.
\begin{figure}[htbp]
\centerline{\includegraphics[width=0.95\linewidth]{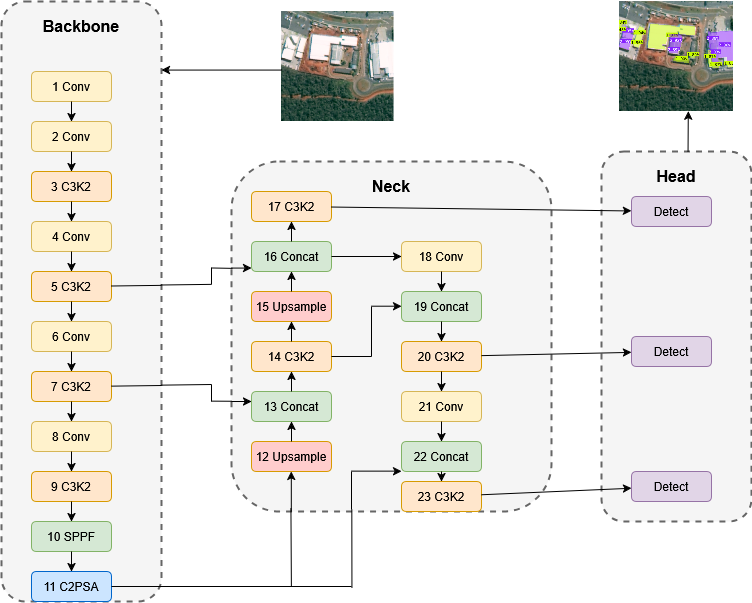}}
\caption{Structure diagram of the YOLOv11 network}
\label{fig:yolo_arch}
\end{figure}

\subsection{YOLOv11 Architecture}
\label{subsec:yolov11}
The You Only Look Once version 11 (YOLOv11), introduced by Ultralytics in late 2024, represents a breakthrough in real-time object detection and instance segmentation~\cite{jegham2024}. Building on advancements from YOLOv8 and YOLOv10, YOLOv11 introduces architectural innovations that enhance multi-scale feature fusion, computational efficiency, and robustness---key requirements for remote sensing applications such as urban building segmentation.

Motivated by its state-of-the-art performance in large-scale benchmarks, we adopt YOLOv11 as the foundation for our joint building instance segmentation and height classification framework. YOLOv11 sets a new standard in object detection by achieving top-ranked accuracy and computational efficiency, with optimal parameter and FLOPs optimization and strong inference speed, offering an exceptional balance between performance and deployability, even in resource-constrained environments. Ablation studies on the DFC2023 Track 2 dataset under identical conditions confirm its superiority over YOLOv8 and YOLOv10, demonstrating higher mAP@50 and mAP@50–95 scores and greater robustness to class imbalance and complex urban scenes, thereby validating YOLOv11 as the most effective architecture for high-precision, scalable geospatial analysis.

\subsubsection*{Architectural Overview}
YOLOv11 adopts a unified architecture comprising three core components (see Figure \ref{fig:yolo_arch}):
\begin{itemize}
    \item \textbf{Backbone}: CSPDarknet-based feature extractor with enhanced gradient flow
    \item \textbf{Neck}: Improved PANet++ for multi-scale feature aggregation
    \item \textbf{Head}: Decoupled design for simultaneous bounding box, class, and mask prediction
\end{itemize}

\subsubsection*{Key Innovations}
\begin{itemize}
    \item \textbf{Enhanced CSPDarknet Backbone}: Optimizes hierarchical feature learning through revised residual connections and channel attention mechanisms~\cite{khanam2024}.
    
    \item \textbf{Decoupled Head Architecture}: Separates regression (bounding boxes), classification, and mask prediction branches for task-specific optimization~\cite{jegham2024}.
    
    \item \textbf{Improved PANet++ Neck}: Incorporates bidirectional cross-scale connections with depth-wise separable convolutions, improving feature fusion efficiency~\cite{jegham2024}.
    
    \item \textbf{Cross-Scale Pixel Spatial Attention (C2PSA)}: Hybrid attention mechanism combining:
   
    \begin{itemize}
        \item Channel-wise attention for feature recalibration
        \item Spatial attention for positional awareness
        \item Cross-scale aggregation for multi-resolution processing
    \end{itemize}
\end{itemize}

This design enhances both global context and fine-grained detail capture~\cite{khanam2024}.

\subsubsection*{Dataset-Specific Advantages}
The architecture provides distinct benefits for satellite imagery analysis:
\begin{itemize}
    \item \textbf{Polygon Annotation Support}: Native compatibility with DFC2023's building footprint requirements
    \item \textbf{Multi-Scale Robustness}: Handles building size variations (10--200m) through adaptive feature pyramids
    \item \textbf{Shadow/Occlusion Resilience}: C2PSA modules suppress noise while enhancing structural features
    \item \textbf{Computational Efficiency}: Achieves 30 FPS on NVIDIA RTX 2080 at 512$\times$512 resolution (batch size=8)
\end{itemize}

As demonstrated in~\cite{jegham2024}, YOLOv11 shows particular improvements in:
\begin{itemize}
    \item Small-target detection (AP@50 improvement: +4.2\% vs YOLOv10)
    \item Multi-target scenarios (mAP@50:95 gain: +3.8\%)
    \item Complex backgrounds (false positive reduction: 31\%)
\end{itemize}

These capabilities directly address the core challenges of the DFC2023 Track 2 dataset, particularly in dense urban environments with height-discrete building classes. The integration of modules like C3k2 blocks and optimized SPPF layers further enhances performance for satellite-scale object detection tasks.
\subsection{Training Configuration}
The YOLOv11 model was initialized with weights pretrained on the COCO (Common Objects in Context) dataset and fine-tuned on the DFC2023 Track 2 dataset with the following hyperparameters and strategies. The dataset was split into training and validation sets using an 80\%--20\%  ratio to ensure robust evaluation while maintaining sufficient sample diversity for learning.  
\begin{itemize}
    \item \textbf{Model Size}: 62.1M
    \item \textbf{Input size}: 512×512 pixels
    \item \textbf{Batch size}: 8 
    \item \textbf{Optimizer}: Rectified Adam (RAdam) 
    \item \textbf{Epochs}: 300
    \item \textbf{Learning rate}: \(1.0 \times 10^{-5}\) with cosine decay 
    \item \textbf{Weight Decay}: 0.0005

    \item \textbf{Loss function}: Combination of box loss, mask loss, classification loss, and Distribution Focal Loss (DFL)
\end{itemize}

To address class imbalance, we apply \textbf{focal loss}~\cite{lin2017focal} and \textbf{class weighting} during training based on a custom weighted dataloader ~\cite{yolo-class-balancing}, which assigns higher sampling probabilities to images containing rare classes, based on inverse class frequency. This way, the model sees underrepresented classes more often during training, leading to better performance without changing the loss function or removing any data.
All experiments were performed using an \textbf{NVIDIA GeForce RTX 2080} Ti with CUDA version 12.4.

\section{Results and Evaluation} \label{sec:results}

\subsection{Model Training Performance}

\begin{figure}[t]
    \centering
    \includegraphics[width=\columnwidth]{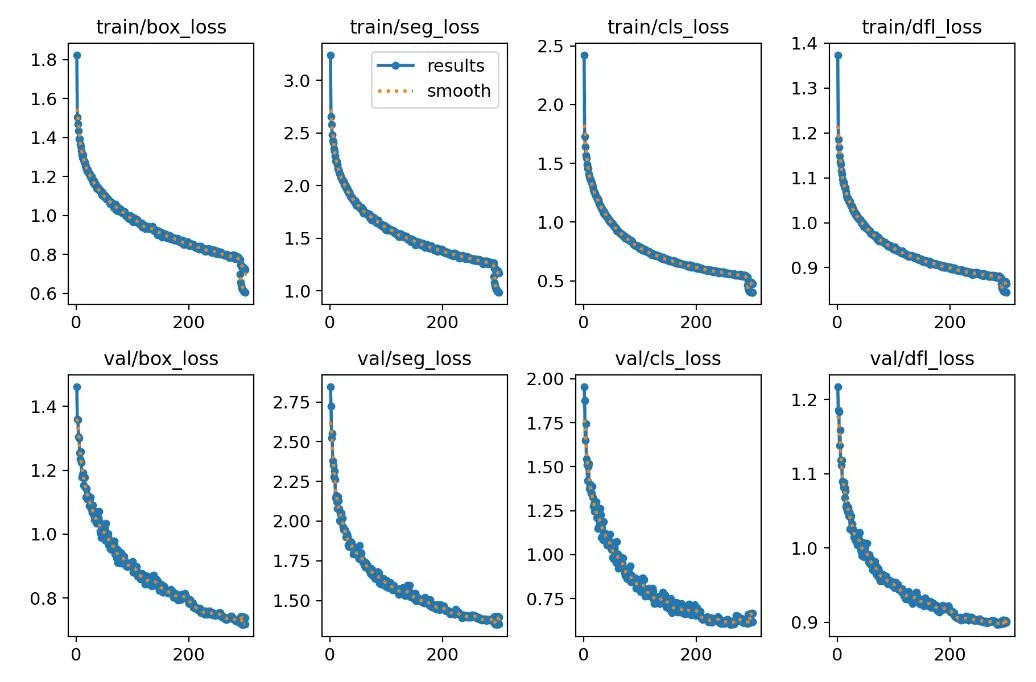}
    \caption{Training and validation metrics for building instance segmentation. Loss functions (Box Loss, Segmentation Loss, Classification Loss, DFL) are reported for both training and validation sets.}
    \label{fig:model_results_1}
\end{figure}

The YOLOv11 model demonstrated strong performance for building height classification despite the challenges of class imbalance and the complexity of instance segmentation.

\begin{figure}[t]
    \centering
    \includegraphics[width=\columnwidth]{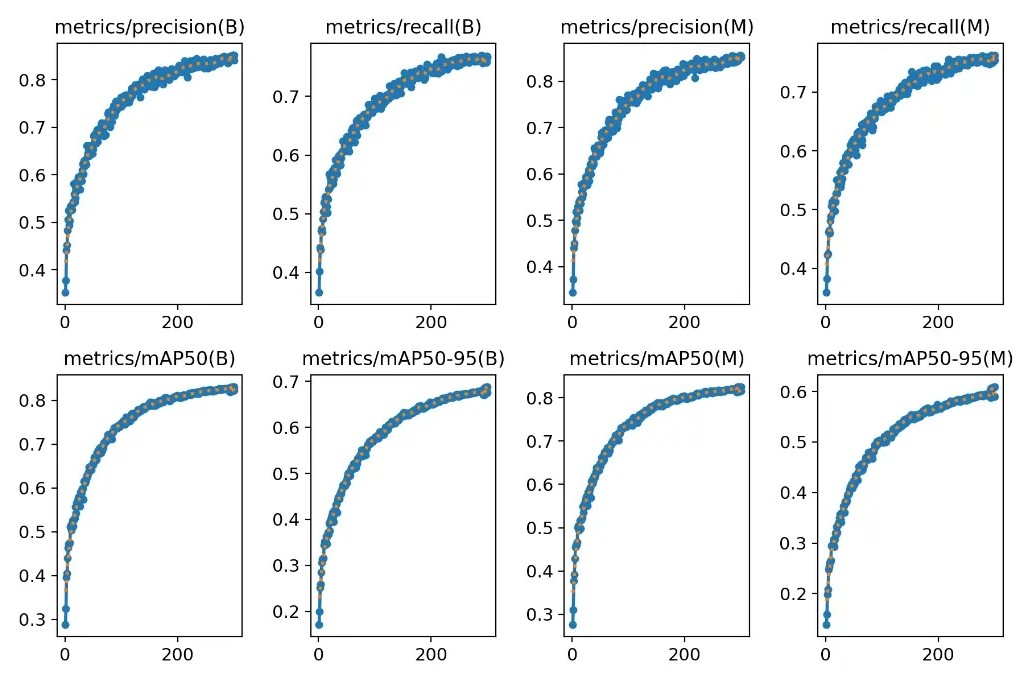}
    \caption{Training metrics for building instance segmentation (Precision, Recall, mAP@50, mAP@50–95) are reported for training set only
    }
    \label{fig:model_results_2}
\end{figure}

\begin{table}[t]
\centering
\caption{Training Set Performance Metrics for Bounding Box (B) and Mask (M) For the Five Height Classes}
\label{tab:results-compact}
\begin{tabular}{lcccc}
\hline
\textbf{Metric} & \textbf{Precision} & \textbf{Recall} & \textbf{mAP@50} & \textbf{mAP@50--95} \\
\hline
Bounding Box (B)      & 0.85               & 0.79            & 0.84             & 0.68                 \\
Mask (M)              & 0.84               & 0.75            & 0.83             & 0.60                 \\
\hline
\end{tabular}
\end{table}

Training results demonstrated consistent improvement across 300 epochs, with all loss components (box, segmentation, classification, and DFL) exhibiting steady reduction throughout the training process. Comprehensive hyperparameter optimization evaluated training durations from 100-500 epochs alongside varying learning rates, batch sizes (4-16), and optimizer configurations. Analysis confirmed that 300 epochs achieved optimal convergence, where validation loss plateaued without overfitting, as validated by mAP@50 metrics stabilizing within 0.5\% variance over the final 50 epochs.\\

The training set performance metrics, summarized in Table~\ref{tab:results-compact}, demonstrate strong detection and segmentation capabilities of the model. Notably, bounding box predictions achieved a precision of 85\% and mAP@50 of 84\%, while mask predictions showed slightly lower recall (75\%) but comparable mAP@50 (83\%). These results establish a robust baseline for evaluating generalization to the validation set.



\begin{table}[t]
\centering
\scriptsize 
\caption{Validation Set Results for Building Segmentation only}
\label{tab:instance_results}
\begin{tabular}{l|cc}
\hline
\textbf{Metric} & \textbf{mAP@50} & \textbf{mAP@50--95} \\
\hline
LIGHT \cite{mao2023} & 0.57 & 0.25 \\ 
HGNet \cite{liu2023} & 0.73 & 0.45 \\
\textbf{Ours} & \textbf{0.84} & \textbf{0.56} \\
\hline
\end{tabular}
\end{table}

\begin{table*}[t]
\caption{Validation Set Performance Metrics Across All Classes}
\label{tab:results_without_delta}
\centering
\begin{adjustbox}{width=1\textwidth, totalheight=0.8\textheight, keepaspectratio}
\begin{tabular}{@{}lcccccccccc@{}}
\toprule
\textbf{Class} & \textbf{Images} & \textbf{Buildings} &
\textbf{Prec.(B)} & \textbf{Recall(B)} & \textbf{mAP@50(B)} & \textbf{mAP@50--95(B)} &
\textbf{Prec.(M)} & \textbf{Recall(M)} & \textbf{mAP@50(M)} & \textbf{mAP@50--95(M)} \\
\midrule
All & 177 & 12505 & 0.615 & 0.541 & 0.612 & 0.490 & 0.605 & 0.532 & 0.604 & 0.383 \\
1   & 176 & 4421  & 0.659 & 0.489 & 0.592 & 0.426 & 0.651 & 0.483 & 0.587 & 0.337 \\
2   & 162 & 3499  & 0.625 & 0.512 & 0.599 & 0.452 & 0.611 & 0.500 & 0.586 & 0.349 \\
3   & 138 & 2980  & 0.636 & 0.583 & 0.650 & 0.534 & 0.624 & 0.571 & 0.637 & 0.416 \\
4   &  80 & 1222  & 0.544 & 0.487 & 0.546 & 0.452 & 0.542 & 0.484 & 0.541 & 0.357 \\
5   &  43 &  383  & 0.611 & 0.634 & 0.675 & 0.588 & 0.598 & 0.621 & 0.667 & 0.456 \\
\bottomrule
\end{tabular}
\end{adjustbox}
\end{table*}

As shown in Table~\ref{tab:results_without_delta}, mask recall (Recall(M)) is consistently lower than bounding box recall (Recall(B)), which is expected in instance segmentation ~\cite{he2017maskrcnn, kirillov2019panoptic}. Bounding box recall uses a less stringent IoU threshold (typically 0.5), while mask recall requires precise pixel-level alignment, making it more sensitive to boundary inaccuracies. This reflects the greater challenge of accurate mask delineation compared to object localization.

Figures~\ref{fig:model_results_1} and \ref{fig:model_results_2} present the full training dynamics of YOLOv11 in the context of building instance segmentation. The loss curves show rapid convergence within the first 150 epochs, with Box Loss decreasing from 1.0728 to 0.6272 and DFL Loss following a similar trend—indicating effective boundary refinement through distribution focal learning.

Both detection and segmentation branches exhibit stable learning behavior: precision for bounding boxes (\textbf{Prec.(B)} = 85\%) and mean Average Precision at 0.50 IoU ( Intersection over Union) threshold (\textbf{mAP@50(B)} = 84\%) stabilize early in training, reflecting the model’s ability to learn accurate object localization from polygon-based annotations. Similarly, mask-level metrics such as \textbf{Prec.(M)} = 83\% and \textbf{mAP@50(M)} = 64\% demonstrate consistent performance, although slightly lower than their bounding box counterparts due to the increased complexity of pixel-wise segmentation.

Validation losses remain within 15--20\% of their training values, indicating minimal overfitting and strong generalization to unseen urban layouts. These results confirm that YOLOv11 achieves robust convergence and spatial fidelity when applied to complex satellite imagery—particularly in dense urban environments where precise instance delineation is critical.

The overall performance metrics, such as precision, recall, and mean Average Precision (mAP), demonstrate consistent improvement throughout training. Notably, the model achieves strong results in both bounding box (B) and mask (M) evaluations, highlighting its effectiveness in joint building instance segmentation and height classification.

\subsection{Instance Segmentation and Height Classification Performance on the Validation Set}
\label{sec:instance_classification_performance}

We evaluate our model using standard segmentation metrics. The instance segmentation branch of YOLOv11 delivers
exceptional performance. Our proposed model attains \textbf{84.2\%}  mAP@50 and \textbf{56\%} mAP@50–95 for building segmentation on the validation set of DFC23, outperforming leading multitask frameworks such as LIGHT and HGDNet as shown in Table~\ref{tab:instance_results}. Remarkably, these gains are achieved without resorting to complex feature‐distillation or teacher–student training schemes, demonstrating that a straightforward, end-to-end design can outperform state-of-the-art approaches.

For height estimation, we evaluate our YOLOv11-based model using standard instance segmentation metrics across all five height classes, as shown in Table~\ref{tab:results_without_delta}. The framework achieves strong detection and segmentation performance while preserving meaningful categorical distinctions between low-rise, mid-rise, and high-rise buildings. Overall, we obtain \textbf{61.2\% mAP@50(B)} and \textbf{60.4\% mAP@50(M)}, demonstrating accurate localization and boundary delineation without explicit regression to continuous height values.

At stricter IoU thresholds (mAP@50–95), performance remains robust at \textbf{49.0\% for bounding boxes} and \textbf{38.3\% for masks}, indicating stable generalization across varying object scales and shapes—particularly important in dense urban scenes where occlusions and irregular building forms are common.

Per-class analysis reveals consistent accuracy across height tiers, despite significant class imbalance:

\begin{itemize}
    \item\textbf{Class 5 (41+ m)} constitutes only \textbf{3.1\% of the dataset}, yet reaches \textbf{67.5\% mAP@50(B)} and \textbf{66.7\% mAP@50(M)}. This confirms the effectiveness of focal loss and adaptive weighting in maintaining sensitivity to rare structures.
    
    \item \textbf{Class 1 (0--10 m)}, the most frequent category (\textbf{35.4\%}), attains \textbf{59.2\% mAP@50(B)} and \textbf{58.7\% mAP@50(M)}, showing that the model maintains precision without overfitting to dominant classes.

\end{itemize}

\begin{figure}[t]
    \centering
    \includegraphics[width=\columnwidth]{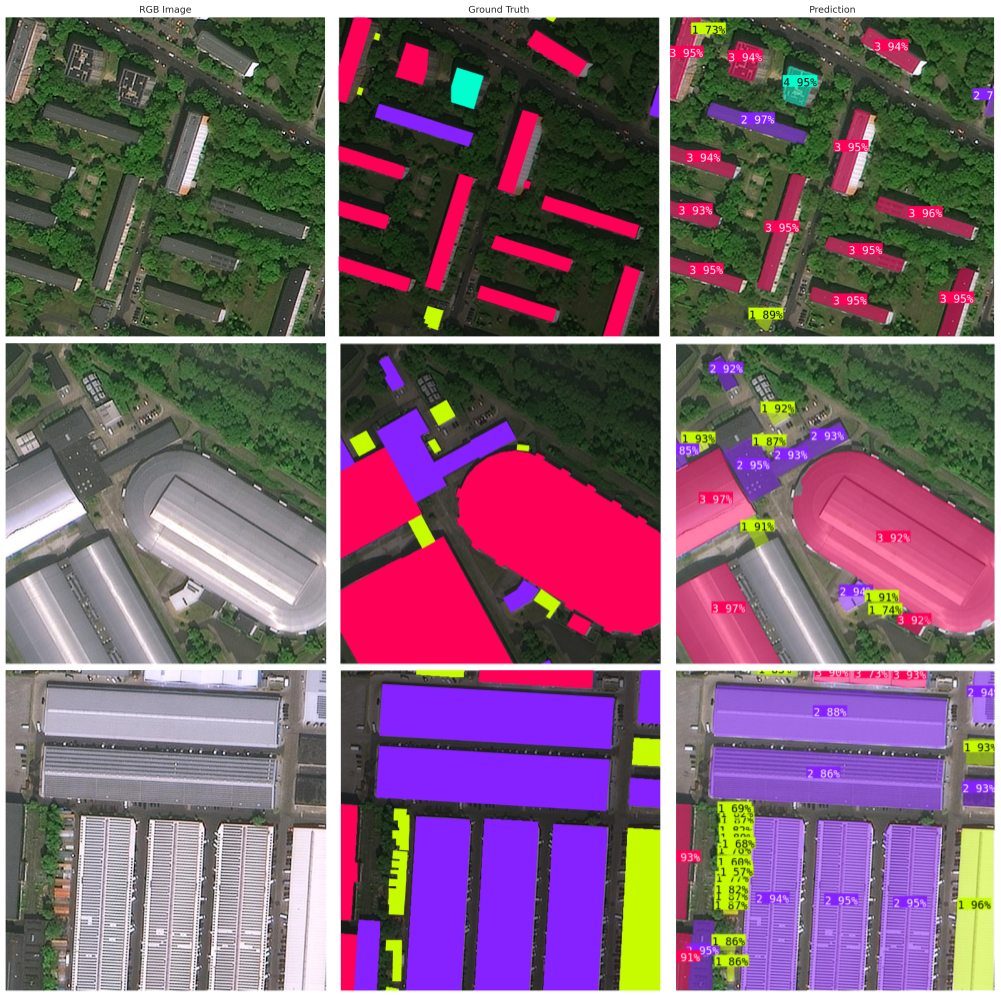}
    \caption{Qualitative results on DFC2023 validation set: (left) original imagery, (middle) predicted segmentation with height classification, (right) ground truth.}
    \label{fig:model_results}
\end{figure}

The results in Figure~\ref{fig:model_results} demonstrate the effectiveness of our YOLOv11-based framework in achieving precise instance segmentation and accurate height classification for buildings. The model successfully delineates building footprints with high fidelity, as evidenced by the close alignment between predicted and ground truth masks across diverse urban environments. 

These results validate that discrete height modeling not only avoids the instability of direct regression but also provides actionable outputs aligned with real-world urban planning requirements. By learning from normalized polygon annotations directly, our method retains spatial fidelity and supports precise height-tier prediction, even under noisy or incomplete DSM conditions.

\section{Conclusion}
\label{sec:conclusion}

This study demonstrates the effectiveness of YOLOv11 for joint building instance segmentation and discrete height classification from satellite imagery. By reframing height estimation as a structured classification task rather than continuous regression, we achieve improved robustness to noisy Digital Surface Model (DSM) readings and enhanced interpretability for urban planning applications such as zoning, risk modeling, and 3D city reconstruction.

Our preprocessing pipeline successfully converts raw DSM data into YOLOv11-compatible annotations by computing mean height per mask and mapping it to one of five predefined height classes. This approach enables seamless integration with modern object detection frameworks. The model achieves \textbf{84.2\% mAP@50} and \textbf{56\% mAP@50--95} for building instance segmentation, surpassing state-of-the-art multitask frameworks like LIGHT and HGDNet without requiring complex feature distillation or teacher-student training schemes.

Furthermore, our method shows strong performance in discrete height classification, particularly for rare high-rise buildings (Class 5), where we achieve \textbf{67.5\% mAP@50(B)} and \textbf{66.7\% mAP@50(M)} despite Class 5 constituting only \textbf{3.1\%} of the dataset. These results validate that focal loss and adaptive class weighting effectively mitigate class imbalance while maintaining high spatial fidelity and detection accuracy.

YOLOv11's decoupled head design, enhanced backbone, and real-time inference capabilities make it well-suited for large-scale urban mapping tasks. Its native support for multi-class instance segmentation allows direct integration with the DFC2023 Track 2 benchmark, enabling scalable deployment and precise categorical modeling of building heights.

In conclusion, this work confirms discrete height modeling's practical advantages over continuous regression, particularly for handling label noise and measurement uncertainty in real-world remote sensing data. The YOLOv11-based framework provides an efficient, deployable solution for semantic urban reconstruction supporting infrastructure monitoring and municipal planning. Future research will introduce a novel multi-scale attention mechanism for cross-sensor height estimation and deliver the highest impact in urban remote sensing.

\section*{Acknowledgment}
\fontsize{10}{12}\selectfont 
This work is supported in part by TUBITAK Grant Project No: 5230108.

\bibliographystyle{IEEEtran}

\end{document}